\documentclass[11pt]{article}
\usepackage[round]{natbib}

\usepackage{graphicx} 
\usepackage{psfrag}
\usepackage{amsmath,amssymb}
\usepackage{amsthm}
\usepackage{verbatim}
\usepackage[usenames,dvipsnames]{color}

\usepackage{epstopdf}

\usepackage{url}
\usepackage{hyperref}

\newtheorem{theorem}{\bf Proposition}

\newtheorem{definition}{\bf Definition}

\newcommand{\bqn}{\begin{eqnarray}}
\newcommand{\eqn}{\end{eqnarray}}
\newcommand{\bq}{\begin{eqnarray*}}
\newcommand{\eq}{\end{eqnarray*}}

\usepackage{color}
\usepackage{hyperref}
\hypersetup{colorlinks,%
citecolor=black,%
filecolor=red,%
linkcolor=red,%
urlcolor=red,%
pdftex}

\newcommand{\red}[1]{\textcolor{red}{#1}}

\begin{document}

\title{Mapping Heritability of Large-Scale Brain Networks with a Billion Connections {\em via} Persistent Homology}

\author{Moo K. Chung$^1$,\;Victoria Vilalta-Gil$^2$, \; Paul J. Rathouz$^1$,\\
 \; Benjamin B. Lahey$^3$, \; David H. Zald$^2$\\
 \\
$^1$University of Wisconsin-Madison,\; $^2$Vanderbilt University, \\
 $^3$University of Chicago\\
 \\
{\tt email:}\red{\tt mkchung@wisc.edu}\\
}

\maketitle

\begin{abstract}
In many human brain network studies, we do not have sufficient number ($n$) of images relative to the number ($p$) of voxels due to the prohibitively expensive cost of scanning enough subjects. Thus, brain network models usually suffer the {\em small-n large-p problem}. Such a problem is often remedied by sparse network models, which are usually solved numerically by optimizing $L1$-penalties. Unfortunately, due to the computational bottleneck associated with optimizing $L1$-penalties, it is not practical to apply such methods to construct large-scale brain networks at the voxel-level. In this paper, we propose a new scalable sparse network model using {\em cross-correlations} that bypass the computational bottleneck. Our model can build sparse brain networks at the voxel level with $p > 25000$. Instead of using a single sparse parameter that may not be optimal in other studies and datasets, the computational speed gain enables us to analyze the collection of networks at every possible sparse parameter in a coherent mathematical framework via {\em persistent homology}. The method is subsequently applied in determining the extent of heritability on a functional brain network at the voxel-level for the first time using twin fMRI.
\end{abstract}

\section{Introduction}

{\em Large-scale brain networks.} In brain imaging, there have been many  attempts to identify high-dimensional imaging features via multivariate approaches including network features \citep{chung.2013.MICCAI,lerch.2006,he.2007,worsley.2005.neural,rao.2008,cao.1999.correlation,he.2008}. Specifically, when the number of voxels (often denoted as $p$) are substantially larger than the number of images (often denoted as $n$), it produces an under-determined model with infinitely many possible solutions. The small-$n$ large-$p$ problem is often remedied by regularizing  the under-determined system with additional sparse penalties. Popular sparse models include sparse correlations \citep{lee.2011.tmi,chung.2013.MICCAI, chung.2015.TMI}, LASSO \citep{bickel.2008,peng.2009,huang.2009,chung.2013.MICCAI},
sparse canonical correlations \citep{avants.2010} and graphical-LASSO \citep{banerjee.2006,banerjee.2008,friedman.2008,huang.2009,huang.2010,mazumder.2012,witten.2011}. Most of these sparse models require optimizing $L1$-norm penalties, which has been the major computational bottleneck for solving large-scale problems in brain imaging. Thus, almost all sparse brain network models in brain imaging have been restricted to a few hundreds nodes or less. As far as we are aware, 2527 MRI features used in {a LASSO model} for Alzheimer's disease \citep{xin.2015} is probably the largest number of features used in any sparse model {in the brain imaging literature}.

In this paper, we propose a new scalable large-scale sparse network model ($p >$ 25000) that yields greater computational speed and efficiency by bypassing the computational bottleneck of optimizing $L1$-penalties. There are few previous studies at speeding up the computation for sparse models. By identifying block diagonal structures in the estimated (inverse) covariance matrix, it is possible to reduce the computational burden in the penalized log-likelihood method \citep{mazumder.2012,witten.2011}.  However, the proposed method substantially differs from \citet{mazumder.2012} and \citet{witten.2011} in that we do not need to assume that the data to follow Gaussianness. Subsequently, there is no need to specify the likelihood function.  Further, the cost functions we are optimizing are different. Specifically, we propose a novel sparse network model based on {\em cross-correlations}. Although cross-correlations are often used in sciences in connection to times series and stochastic processes \citep{worsley.2005.neural,worsley.2005.royal}, the sparse version of cross-correlation has been somewhat neglected. \\

{\em Persistent network homology.} Any sparse model $\mathcal{G}(\lambda)$ is usually parameterized by a tuning parameter $\lambda$ that controls the sparsity of the representation.  Increasing the sparse parameter makes the  solution more sparse. Sparse models are inherently multiscale, where the scale of the models is determined by the sparse parameter. However, many existing sparse network approaches use a fixed parameter $\lambda$ that may not be optimal in other datasets or studies \citep{chung.2013.MICCAI,lee.2012.tmi}. Depending on the choice of the sparse parameter, the final classification and statistical results will be totally different \citep{lee.2012.tmi,chung.2013.MICCAI,chung.2015.TMI}. Thus, there is a need to develop a multiscale network analysis framework that provides a consistent statistical results and interpretation regardless of the choice of parameter. Persistent homology, a branch of algebraic topology, offers one possible solution to the multiscale problem \citep{carlsson.2008,edelsbrunner.2008}. Instead of looking at images and networks at a fixed scale, as usually done in traditional analysis approaches, persistent homology observes the changes of topological features over different scales and finds the most persistent topological features that are robust under noise perturbations. This robust performance under different scales is needed for most network models that are parameter and scale dependent.

In this paper, instead of building networks at one fixed sparse parameter that may not be optimal, we propose to build the collection of sparse models over every possible sparse parameter using a {\em graph filtration}, a persistent homological construct  \citep{chung.2013.MICCAI,lee.2012.tmi}. The graph filtration is a threshold-free framework for analyzing a family of graphs but requires hierarchically building specific nested subgraph structures. The proposed method share some similarities to the existing multi-thresholding or multi-resolution network models  that use many different arbitrary thresholds or scales  \citep{achard.2006,he.2008,kim.2015,lee.2012.tmi,supekar.2008}. However, such approaches have not often applied to sparse models even in an ad-hoc fashion before. Additionally, building sparse models for multiple sparse parameters can causes an additional computational bottleneck to the already computationally demanding problem  as well as introducing an additional problem of choosing optimal parameters. {The proposed persistent homological approach} can address all these shortcomings within a unified mathematical framework. \\

{\em Heritability of brain networks.} Many brain imaging studies have shown the widespread heritability of neuroanatomical structures in magnetic resonance imaging (MRI) \citep{mckay.2014} and diffusion tensor imaging (DTI) \citep{chiang.2011}. However, these structural imaging studies use univariate imaging phenotypes such as cortical thickness and fractional anisotropy (FA)  in determining heritability at each voxel or regions of interest. There are also few fMRI and EEG studies that use functional activations as possible imaging phenotypes at each voxel \citep{blokland.2011,glahn.2010,smit.2008}. Compared to many existing studies on univariate phenotypes, there are not many  studies on the heritability of brain networks \citep{blokland.2011}. Measures of network topology and features may be worth investigating as intermediate phenotypes or endophenotypes, that indicate the genetic risk for a neuropsychiatric disorder \citep{bullmore.2009}. However, the brain network analysis has not yet been adapted for this purpose. Determining the extent of  heritability of brain networks with large number of nodes is the first necessary prerequisite for identifying network-based endophenotypes. This requires constructing the large-scale brain networks by taking every voxel in the brain as network nodes with at least {a billion connections}, which is a serious computational burden. 


Motivated by these neuroscientific and methodological needs, we propose to develop algorithms for building large-scale brain networks based on {\em sparse cross-correlations}. The proposed sparse network model is used in constructing large-scale brain networks at the voxel level in fMRI consisting of monozygotic (MZ) and dizygotic (DZ) twin pairs. The twin study design in brain imaging offers a very effective way of determining {the heritability of the multiple features of the human brain}. Heritability (broad sense) can be measured by comparing the resemblance among monozygotic (MZ) twins and dizygotic (DZ) twins using Falconer's method \citep{falconer.1995,tenesa.2013} at the voxel-level. 
As a way of quantifying the genetic contribution of phenotypes, the heritability index (HI) has been extensively used in twin studies. Heritability index (HI) is the the amount of genetic variations in a phenotype. For instance, 60\% HI implies the phenotype is 60\% heritable and 40\% environmental. 
Except for few well known neural circuits \citep{glahn.2010,van.2013},  the extent to which {heritability influences functional brain networks} at the voxel-level is not {well established}. Although HI is a well formulated concept, it is unclear {how to best apply} HI to networks. In this paper, we generalize the voxel-wise univariate {\em heritability index} (HI) into a network-level multivariate feature called {\em heritability graph index} (HGI), which is subsequently used in determining the genetic effects at the network-level. The proposed framework is used  to determine the first-ever voxel-level heritability map of functional brain network. \\



\section{Methods}
\subsection{Sparse Cross-Correlations}
Let $V= \{ v_1, \cdots, v_p \}$ be a node set where data is observed. We expect the number of nodes $p$ to be significantly larger than the number of images $n$, i.e., $p \gg n$. In this study,  we have $p=$ 25972 voxels in the brain that serves as the node set. Let $x_k(v_i)$ and $y_k(v_i)$ be the $k$-th paired scalar measurements at node $v_i$. Denote ${\bf x}(v_i) =(x_1(v_i), \cdots, x_n(v_i))'$ and ${\bf y}(v_i) =(y_1(v_i), \cdots, y_n(v_i))'$ be the paired data vectors over $n$ different images at voxel $v_i$.  Center and scale ${\bf x}$ and ${\bf y}$ such that
$$ \sum_{k=1}^n x_k(v_i) = \sum_{k=1}^n y_k (v_i) = 0, $$
$$\| {\bf x}(v_i) \|^2 = {\bf x}'(v_i){\bf x}(v_i) =  \| {\bf y}(v_i) \|^2  = {\bf y}'(v_i){\bf y}(v_i) = 1$$
for all $v_i$. 
The reasons for centering and scaling will soon be obvious. We set up a linear model between ${\bf x}(v_i)$ and ${\bf y}(v_j)$:   
\bqn {\bf y}(v_j)= b_{ij} \; {\bf x}(v_i) + {\bf e}, \label{eq:LRG}\eqn
where ${\bf e}$ is the zero-mean error vector whose components are independent and identically distributed. 
Since the data are all centered, we do not have the intercept in linear regression (\ref{eq:LRG}). The least squares estimation (LSE) of $b_{ij}$ that minimizes the L2-norm 
\bqn \sum_{i,j=1}^p
\parallel {\bf y}(v_j) - b_{ij} \; {\bf x}({v_i}) \parallel^{2} \label{eq:LRG2}\eqn
is given by 
\bqn \widehat{b}_{ij} = {\bf x}'(v_i) {\bf y}(v_j),\label{eq:gamma}\eqn
which are the (sample)  {\em  cross-correlations} \citep{worsley.2005.neural,worsley.2005.royal}.  
 The cross-correlation is invariant under the centering and scaling operations. The sparse version of L2-norm (\ref{eq:LRG2}) is given by

\bqn
\label{eq:lasso_corr}
F(\beta; {\bf x}, {\bf y},\lambda) = \frac{1}{2} \sum_{i,j =1}^p 
\parallel {\bf y}(v_j) - \beta_{ij} \; {\bf x}({v_i}) \parallel^{2} + \lambda \sum_{i,j =1}^p  | \beta_{ij} |.
\eqn
The {\em sparse cross-correlation} is then obtained by minimizing over every possible $\beta_{ij} \in \mathbb{R}$: 
\bqn \widehat \beta (\lambda) = \arg \min_{\beta}  F(\beta; {\bf x}, {\bf y}, \lambda). \label{eq:betamin} \eqn
The estimated sparse cross-correlations $\widehat{\beta}(\lambda) = (\widehat \beta_{ij}(\lambda))$  shrink toward zero as sparse parameter $\lambda  \geq 0$ increases.  
The direct optimization of (\ref{eq:lasso_corr}) for large $p$ is computationally demanding. However, there is no need to optimize (\ref{eq:lasso_corr}) numerically using the coordinate descent learning or the active-set algorithm as often done in sparse optimization \citep{peng.2009,friedman.2008}. We can show that the minimization of (\ref{eq:lasso_corr}) is simply done algebraically.

\begin{theorem}
\label{theorem:SCC} 
For $\lambda \geq 0$, the minimizer of $F(\beta; {\bf x}, {\bf y}, \lambda)$ is given by 
 \bqn \widehat{\beta}_{ij}(\lambda) 
 = \begin{cases} 
 {\bf x}'(v_i) {\bf y}(v_j)  - \lambda & \mbox{ if }   \;  {\bf x}'(v_i) {\bf y}(v_j)  > \lambda \\
 0                                               & \mbox{ if }   \; | {\bf x}'(v_i) {\bf y}(v_j)|  \le \lambda \\ 
 {\bf x}'(v_i) {\bf y}(v_j)  + \lambda & \mbox{ if }   \;  {\bf x}'(v_i) {\bf y}(v_j) < -\lambda
\end{cases}.
\label{eq:cases}
\eqn
\end{theorem}
See Appendix for proof. Although it is not obvious, Proposition \ref{theorem:SCC} is related to the orthogonal design in LASSO  \citep{tibshirani.1996} and the soft-shrinkage in wavelets \citep{donoho.1995}. To see this,  let us transform linear equations (\ref{eq:LRG})
 into a index-free matrix equation:
\bqn
\left[
\begin{array}{ccc}
 {\bf y}(v_1)   & \cdots & {\bf y}(v_1)  \\
{\bf y}(v_2)  & \cdots & {\bf y}(v_2)   \\
  \vdots &     \ddots & \vdots\\  
  {\bf y}(v_p)   & \cdots & {\bf y}(v_p)  
\end{array}
\right]
= \left[
\begin{array}{cccc}
b_{11}{\bf x}(v_1)  &  b_{21}{\bf x}(v_2) & \cdots & b_{p1}{\bf x}(v_p)  \\
b_{12}{\bf x}(v_1)  &  b_{22}{\bf x}(v_2) & \cdots & b_{p2}{\bf x}(v_p)  \\
  \vdots &  \vdots &   \ddots & \vdots\\
b_{1p}{\bf x}(v_1)  &  b_{2p}{\bf x}(v_2) & \cdots & b_{pp}{\bf x}(v_p)  \\  
\end{array}
\right]  + \left[
\begin{array}{ccc}
{\bf e}  &  \cdots & {\bf e}  \\
{\bf e}  &  \cdots & {\bf e}  \\
\vdots & \ddots  &  \vdots \\
{\bf e}  &  \cdots & {\bf e}  
\end{array}
\right].
\label{eq:bigmatrix}
\eqn
Equation (\ref{eq:bigmatrix}) can be vectorized as follows.
\bqn
\left[
\begin{array}{c}
{\bf y}(v_1)\\
  \vdots\\
{\bf y}(v_p)\\
\hline
\vdots\\
\hline
{\bf y}(v_1)\\
  \vdots\\
{\bf y}(v_p)
\end{array}
\right]
= \left[\begin{array}{ccc}
\begin{array}{ccc}
{\bf x}(v_1) & \cdots & 0 \\
\vdots & \ddots & \vdots \\
0 & \cdots & {\bf x}(v_1)
\end{array}
& \cdots &  \mbox{{\huge 0}}  \\
  \vdots & \ddots & \vdots\\
 \mbox{{\huge0}} & \cdots
 &
\begin{array}{ccc}
{\bf x}(v_p) & \cdots & 0 \\
\vdots & \ddots & \vdots \\
0 & \cdots & {\bf x}(v_p)
\end{array}
\end{array}\right]
\left[
\begin{array}{c}
b_{11}\\
\vdots\\
b_{p1}\\
\hline
\vdots\\
\hline
b_{1p}\\
\vdots\\
b_{pp}
\end{array}
\right]
 + 
\left[
\begin{array}{c}
{\bf e}\\
\vdots\\
{\bf e}\\
\hline
\vdots\\
\hline
{\bf e}\\
\vdots\\
{\bf e}
\end{array}
\right].
\label{eq:bigvector}
\eqn
(\ref{eq:bigvector}) can be written in a more compact form. Let
\bq {\bf X}_{n \times p} &=& [{\bf x}(v_1) \; {\bf x}(v_2) \cdots \; {\bf x}(v_p) ]\\
{\bf Y}_{n \times p} &=& [{\bf y}(v_1) \; {\bf y}(v_2) \cdots \; {\bf y}(v_p) ]\\
{\bf 1}_{a \times b} & = & 
\left[\begin{array}{cccc}
1 &1 &  \cdots & 1\\
\vdots & \vdots & \ddots & \vdots\\
1 & 1 & \cdots & 1
\end{array}\right]_{a \times b}.
\eq
Then (\ref{eq:bigvector}) can be written as 
\bqn {\bf 1}_{p \times 1} \otimes vec({\bf Y}) =  \mathbb{X}_{np^2 \times p^2}\; vec(b) + {\bf 1}_{np^2 \times 1} \otimes {\bf e}, \label{eq:gigantic}\eqn
 where {\em vec} is the vectorization operation. The block diagonal design matrix $\mathbb{X}$ consists of $p$ diagonal blocks $I_p \otimes {\bf x}(v_1), \cdots, I_p \otimes {\bf x}(v_p)$, where $I_{p}$ is $p \times p$ identity matrix. Subseqeuntly, 
 $\mathbb{X}' \mathbb{X}$ is again a block diagonal matrix, where the $i$-th block is
 $$[I_p \otimes {\bf x}(v_i)]' [I_p \otimes {\bf x}(v_i)] = I_p \otimes [{\bf x}(v_i)' {\bf x}(v_i)] = I_p.$$
Thus, $\mathbb{X}$ is an orthogonal design. However, our formulation is {\em not} exactly the orthogonal design of LASSO as specified in  \cite{tibshirani.1996} since the noise components in (\ref{eq:gigantic}) are not independent. Further in standard LASSO, there are more columns than rows in  $\mathbb{X}$. In our case, there are $n$ times more rows.

\begin{figure}[h]
\centering
\includegraphics[width=1\linewidth]{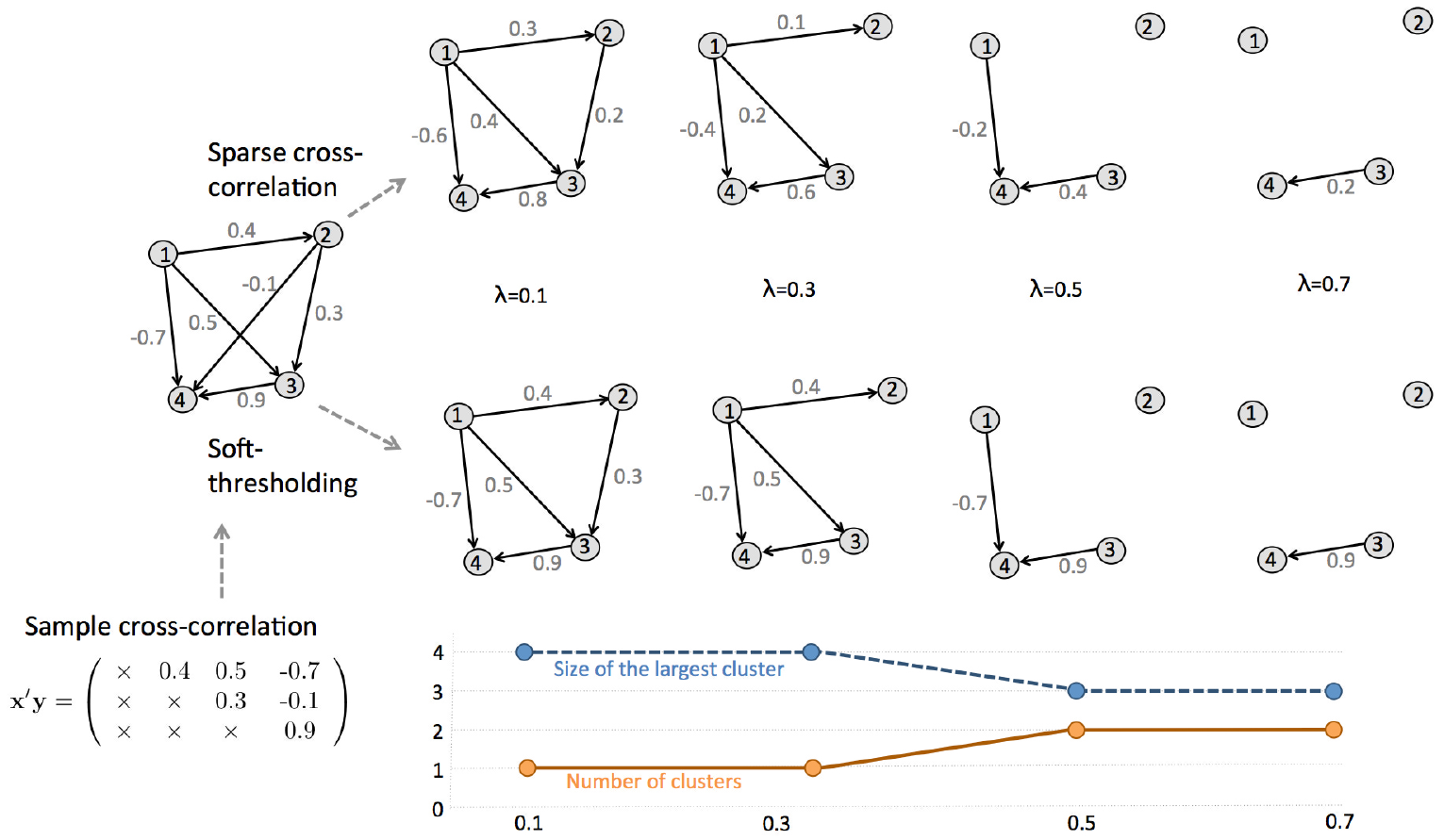} 
\caption{
Top: The sparse cross-correlations are estimated  by minimizing the L1 cost function (\ref{eq:lasso_corr}) for 4 different sparse parameters $\lambda$ (Proposition \ref{theorem:SCC}). The edge weights shrinked to zero are removed. Bottom: the equivalent binary graph can be obtained by the simple soft-thresholding rule (Proposition \ref{theorem:maximal}), i.e., simply thresholding the sample cross-correlations at $\lambda$. The number of clusters (denoted as $\#$) and the size of the largest cluster (denoted as $\&$) are displayed over different $\lambda$ values. 
}
\label{fig:HI-illustration}
\end{figure}

Proposition \ref{theorem:SCC} generalizes the sparse correlation case given in \citep{chung.2013.MICCAI}. 
Figure \ref{fig:HI-illustration}-top displays an example of obtaining sparse cross-correlations from the initial sample cross-correlation matrix
$${\bf X}'{\bf Y}=
\left(
\begin{array}{cccc}
 \times & \mbox{0.4}  & \mbox{0.5} & \mbox{ -0.7} \\
 \times&  \times &  \mbox{0.3} & \mbox{ -0.1} \\
   \times &   \times &  \times  & \mbox{ 0.9}
\end{array}
\right)
$$
using Proposition \ref{theorem:SCC}. For simplicity, only the upper triangle part of the sample cross-correlation is demonstrated.

\subsection{Graph Filtrations}
From estimated sparse cross-correlations $\widehat{\beta}(\lambda)$, we will first build the graph representation for subsequent brain network quantification. Instead of trying to determine the optimal parameter $\lambda$ and fix $\lambda$ as often done in many sparse network models \citep{peng.2009,huang.2009,xin.2015}, we will analyze the sparse networks over every possible $\lambda$. {A single optimal parameter} may not be universally optimal across different datasets or even sufficient.  This new approach enables us to build a multiscale network framework over sparse parameter $\lambda$.

\begin{definition}
\label{def:notations}
Suppose weighted graph $G=(V, W)$ is formed by the pair of node set  $V=\{v_1, \cdots, v_p \}$ and edge weights $W_{p \times p}= (w_{ij})$. Let $^0$ and $^{+\lambda}$ be binary operations on weighted graphs such that $G^0= (V, W^0), W^0 = (w^0_{ij}) $ and $G^{+\lambda} = (V, W^{+\lambda}), W^{+\lambda} = (w^{+\lambda}_{ij})$ with
$$w^0_{ij}=
\begin{cases}
1 &\; \mbox{  if } w_{ij} \neq 0;\\
0 & \; \mbox{ otherwise}
\end{cases}
\;\;\;\;\;\; \mbox{ and }  \;\;\;\;\;\;
w^{+\lambda}_{ij} =
\begin{cases}
1 &\; \mbox{  if } w_{ij} > \lambda;\\
0 & \; \mbox{ otherwise}
\end{cases}.
$$
\end{definition}
The operations $^0$ and $^+$ threshold edge weights and make the weighted graph binary. 

 \begin{definition}
 \label{def:filtration}
 For graphs $G_1 =(V, W), W=(w_{ij})$ and $G_2=(V, Z), Z=(z_{ij})$,
$G_2$ is a subset of  $G_1$, i.e., $G_1 \supset G_2$, 
if $w_{ij} \geq z_{ij}$ for all $i,j$. The collection of nested subgraphs
$$G_1 \supset G_2 \supset \cdots \supset G_k$$
is called a graph filtration. $k$ is the level of the filtration. 
\end{definition}

Definition \ref{def:filtration} extends the concept of  the graph filtration heuristically given in \citet{lee.2012.tmi} and \citet{chung.2013.MICCAI}, where only undirected graphs are considered, to more general directed graphs. 
Graph filtrations is a special case of Rips filtrations  often studied in  persistent homology \citep{carlsson.2008,edelsbrunner.2008,singh.2008,ghrist.2008}. 
From Definitions \ref{def:notations} and \ref{def:filtration}, for arbitrary weighted graph $G=(V,W)$, we have 
\bqn G^{+\lambda_1} \supset G^{+ \lambda_2} \supset G^{+ \lambda_3} \supset \cdots  \label{eq:GF}\eqn
for any $\lambda_1 \leq \lambda_2 \leq \lambda_3 \cdots$. Figure \ref{fig:HI-illustration}-bottom illustrates a graph filtration obtained with $^+$ operation.

Using these concepts, we will build persistent homology on sparse network  $\mathcal{G}(\lambda) = (V, \widehat{\beta}(\lambda))$ with sparse cross-correlations $\widehat{\beta}(\lambda)$ obtained from (\ref{eq:betamin}).  For this, we will first construct the collection of infinitely many binary graphs $\{ \mathcal{G}^0(\lambda): \lambda \in \mathbb{R}^+\}$. 

\begin{theorem} 
\label{theorem:maximal} Let  $\mathcal{G}^0(\lambda)$ be the binary graph obtained from sparse network $\mathcal{G}(\lambda) = (V,\widehat{\beta}(\lambda))$ where $\widehat{\beta}(\lambda)$ are sparse correlations. Consider another graph $\mathcal{H} = (V, \rho)$, where the edge weights $\rho_{ij} =  |{\bf x}(v_i)'{\bf y}(v_j)|$ are the sample cross-correlations. Then, we have the following.\\
(1) Soft-thresholding rule: $\mathcal{G}^0(\lambda) =\mathcal{H}^{+\lambda}$ for all $\lambda \geq 0$.\\
(2) The collection of binary graphs $\{ \mathcal{G}^0(\lambda): \lambda \in \mathbb{R}^+ \}$ 
 forms a graph filtration. \\
(3) Order edge weights $\rho_{ij}$ such that
$$ \rho_{(0)} = 0 \leq \rho_{(1)}=\min_{i,j} \rho_{ij} \leq  \cdots \leq \rho_{(q)}  = \max_{i,j} \rho_{ij}.$$ Then, for any $\lambda \geq 0$, $\mathcal{G}^0(\lambda) = \mathcal{H}^{+\rho_{(i)}}$ for some $i$.
\end{theorem}

See Appendix for proof. Proposition \ref{theorem:maximal}-(1) 
enables us to construct large-scale sparse networks by the simple {\em soft thresholding} rule. This completely bypasses numerical optimizations that have been the main computational bottleneck in the field. Figure \ref{fig:HI-illustration} illustrates the soft-thresholding rule in obtaining the equivalent sparse binary graph without optimization. 

Proposition \ref{theorem:maximal}-(2) and \ref{theorem:maximal}-(3) further show that we can monotonically map the collection of constructed graphs $\{ \mathcal{G}^0(\lambda): \lambda \in \mathbb{R}^+\}$ into finite graph filtration 
\bqn   \mathcal{H}^{+\rho_{(0)}}  \supset \mathcal{H}^{+\rho_{(1)}}   \supset \cdots  \supset  \mathcal{H}^{+\rho_{(q)}}\label{eq:maximal}.\eqn
For a graph with $p$ nodes, the maximum possible number of edges is $(p^2-p)/2$, which is obtained in a complete graph. $(p^2-p)/2 +1$ is the upper limit for the level of filtration in (\ref{eq:maximal}).

\begin{figure}
\centering
\includegraphics[width=1\linewidth]{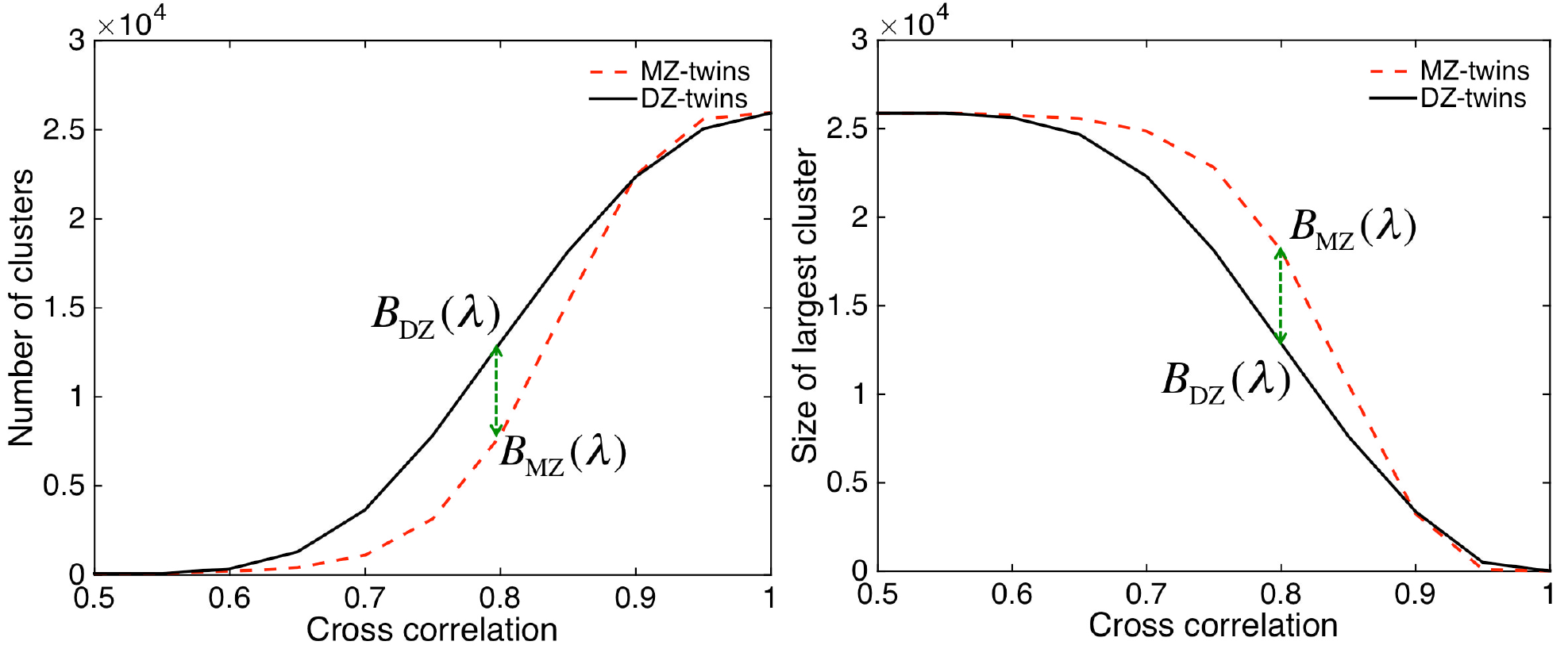}
\caption{The result of graph filtrations on twin fMRI data. The number of disjoint clusters (left) and the size of largest cluster (right) are plotted over  the filtration values. At a given filtration value, MZ-twins have smaller number of clusters but larger cluster size.}
\label{fig:HGIplot}
\end{figure}

\subsection{Statistical Inference on Graph Filtrations}
We propose a new statistical inference procedure for persistent homology. The method is applicable to many filtrations in persistent homology including Rips, Morse as well as graph filtrations. {We apply monotonic graph functions as multiscale features} for statistical inference.

\begin{definition}
Let $\#G$ be the number of connected components (or disjoint clusters) of graph $G$ and $\&G$ be the size of the largest component (or cluster) of $G$. 
\end{definition}
The graph functions $\#$ and $\&$ are monotonic over graph filtrations. For filtration $G_1 \supset G_2 \supset  \cdots$, Note $\#(G_1) \leq \#(G_2) \leq \cdots$ and $\&(G_1) \geq \&(G_2) \geq \cdots$. Figure \ref{fig:HI-illustration} illustrates this monotonicity on the 4-nodes sparse network. 


\begin{theorem} 
\label{theorem:MST}
Let $M$ be the minimum spanning tree (MST) of  weighted graph $G= (V, \rho)$ with edge wights $\rho=(\rho_{ij})$. Suppose the edge weights of $\mathcal{M}$ are sorted as $\rho_{(0)}= -\infty \leq \rho_{(1)} \leq  \cdots \leq \rho_{(p-1)}$. Then for all $\lambda$,
$\#M^{+\rho_{(i)}} = \#G^{+\lambda} \leq \#M^{+\rho_{(i+1)}}$  for some $i$.
\end{theorem}
See Appendix for proof. {We do not show it here} but an analogue statement can be similarly obtained for $\&(G)$. Proposition \ref{theorem:MST} can be used to {\em monotonically} map infinitely many $G^{+\lambda}$ 
to only $p$ number of sorted features
\bqn \# M^{+(-\infty)} \leq \# M^{+\rho_{(1)}}  \leq   \cdots  \leq  \# M^{+\rho_{(p-1)}}. \label{eq:maximal1}\eqn
Similarly, we can monotonically map $G^{+\lambda}$ as
\bqn \& M^{+(-\infty)} \geq \& M^{+\rho_{(1)}}  \geq   \cdots  \geq  \& M^{+\rho_{(p-1)}}. \label{eq:maximal2} \eqn
Figure \ref{fig:HGIplot} displays the plot of the number of clusters  and the size of the largest cluster over filtration values  for our twin fMRI data with $p=25972$ nodes. There are clear group discriminations. The statistical significance for of the discrimination can be computed as follows. 

Let $B_i(\lambda)$ be monotonic graph functions, such as $\#$ and $\&$, over graph filtrations $G_i^{+\lambda}$. Then we are interested in testing the null hypothesis $H_0$ of the equivalence of the two set of monotonic functions: $$H_0: B_1 (\lambda) = B_2 (\lambda) \; \mbox{ for all } \lambda.$$
As a test statistic, we propose to use $$D_p = \sup_{\lambda} | B_1 (\lambda) - B_2 (\lambda)|,$$
which can be viewed as the distance between two graph filtrations.
The test statistic $D_p$ is related to the two-sample Kolmogorove-Smirnov (KS) test statistic \citep{chung.2013.MICCAI,gibbons.1992,bohm.2010}. 
The test statistic takes care of the multiple comparisons over sparse parameters so there is no need to apply additional multiple comparisons correction procedure. The asymptotic probability distribution of $D_{p}$ can be determined. 

\begin{theorem}
\label{theorem:lim}
$\lim_{p \to \infty} P( D_{p}/\sqrt{2(p-1)} \leq d)  = 1 - 2 \sum_{i=1}^{\infty} (-1)^{i-1}e^{-2i^2d^2}.$ 
\end{theorem}
See Appendix for proof. Note Proposition \ref{theorem:lim} is nonparametric test and does not assume any statistical distribution  other than that $B_1$ and $B_2$ are monotonic. From Proposition \ref{theorem:lim}, $p$-value under $H_0$ is computed as
$$\mbox{$p$-value} = 2 e^{-d_{o}^2} - 2e^{-8d_{o}^2} + 2 e^{-18d_{o}^2} \cdots,$$ where $d_{o}$ is the observed value of $D_{p}/\sqrt{2(p-1)}$ in the data.  For any large observed value $d_0 \geq 2$, the second term is in the order of $10^{-14}$ and insignificant. Even for small observed $d_0$, the expansion converges quickly. 

\subsection{Validation Study}
The proposed method was validated in two simulations with the known ground truths. The simulations {were}  performed 1000 times and the average results {were} reported. There are three groups and the sample size is $n=20$ in each group and the number of nodes are $p=100$. In Group 1 and 2, data $x_{k}(v_i)$ at node $v_i$ {was} simulated as standard normal, i.e., $x_{k}(v_i) \sim N(0,1)$. The paired data {was| simulated as $y_k(v_i) = x_k(v_i) + N(0, 0.02^2)$ for all the nodes. The simulated networks can be found in Figure \ref{fig:simul}. Following the proposed method, we obtained the $p$-values of $0.712 \pm 0.331$  and $0.462 \pm 0.413$ for the number of clusters and the size of largest cluster. We could not detect any group differences as expected.

\begin{figure}[h]
\centering
\includegraphics[width=0.9\linewidth]{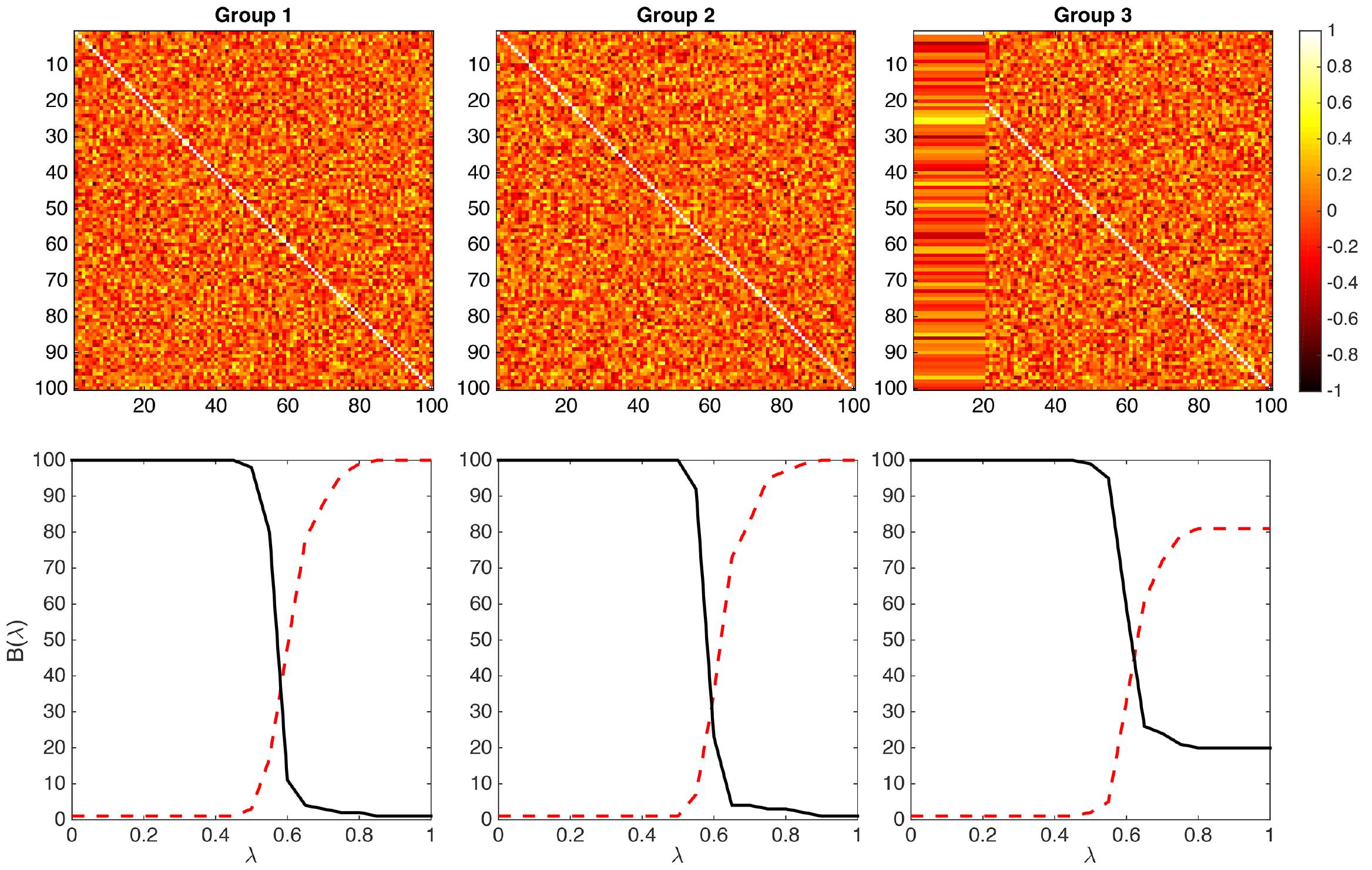}
\caption{Simulation studies. Three groups are simulated. Group 1 and Group 2 are generated independently but identically. The resulting $B(\lambda)$ plots are similar. The dotted red line is the number of clusters and the solid black line is the size of the largest cluster. No statistically significant differences can be found between Groups 1 and 2. Group 3 is generated independently but identically {as} Group 1 but additional dependency is added for the first 10 nodes. The resulting $B(\lambda)$ plots show statistically significant differences. This simulation is repeatedly done 1000 times.} 
\label{fig:simul}
\end{figure}

Group 3 {was} generated identically but independently like Group 1 and 2 but additional dependency {was} added.  For ten nodes indexed by  $j=1, 2, \cdots, 10$, we simulated $y_k(v_j)= x_k(v_1) + N(0, 0.02^2).$ This dependency gives high connection differences between Groups 1 and 3. The simulated network can be found in Figure \ref{fig:simul}. We obtained the $p$-values of $0.025 \pm 0.020$ and $0.004 \pm 0.010$ for the number of clusters and the size of largest cluster demonstrating {that we can} detect the network differences as expected.

  

\section{Application}
\subsection{Twin fMRI Dataset}

The study consists of 11 monozygotic (MZ) and 9 same-sex dizygotic (DZ) twin pairs of 3T functional magnetic resonance images (fMRI) acquired in Intera-Achiava Phillips MRI scanners with a state-of-the-art 32 channel SENSE head coil. Research was approved through the Vanderbilt University Institutional Review Board (IRB). 
Blood oxygenation-level dependent (BOLD) functional images were acquired with a gradient echoplanar sequence, with 38 axial-oblique slices (3 mm thick, 0.3 mm gap), 80 $\times$ 80 in plane resolution, TR (repetition time) 2000 ms, TE (echo time) 25 ms, flip angle = 90.  Subjects completed monetary incentive delay task \citep{knutson.2001} of 3 runs of 40 trials.  A total 120  trials consisting of 40 $\$$0, 40 $\$$1 and 40 $\$$5 rewards were pseudo randomly split into 3 runs. 
The aim of the task is to rapidly respond to a target in order to earn rewards.
Trials begin with a cue indicating the amount of money  that's at stake for that specific trial, then there is a delay period between the cue and the target in which participants prepare for the target to appear. Then a white star (target) flashes rapidly on the screen. If the participants hit the response button while the star is on the screen win the money. If they hit too late, they do not win the money. After the response, and a brief delay a feedback slide is shown indicating to the participants whether they hit the target on time or not and the amount of money they made for that specific trial. 

\begin{figure}[ht]
\centering
\includegraphics[width=0.75\linewidth]{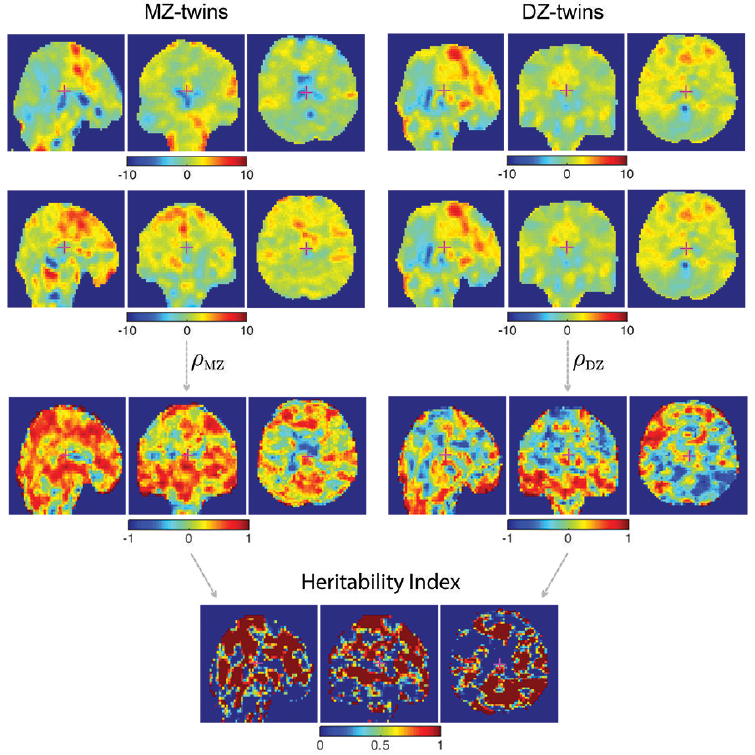}
\caption{Representative MZ- and DZ-twin pairs of the contrast maps obtained from the first level analysis testing the significance of the delay in hitting the response button in \$5 reward in contrast to \$0 reward. Middle: The correlation of the contrast maps within each twin types.  Bottom: The heritability index measures twice the difference between the correlations.}
\label{fig:HI-corr1}
\end{figure}

The fMRI dimensions after preprocessing are 53 $\times$ 63 $\times$ 46 and the voxel size is 3mm cube. There are a total of $p=25972$ voxels in the brain template. Figure \ref{fig:HI-corr1} shows the outline of the template. fMRI data went through spatial normalization to a template following the standard SPM pipeline \citep{penny.2011}. Images 
were processed with  Gaussian kernel smoothing with 8mm full-width-at-half-maximum (FWHM) spatially. Neuroscientifically, we are  interested in knowing {\em the extent of the genetic influence on the functional brain networks} of  these participants while anticipating the high 
reward as measured 
by activity during the delay that occurs between the reward cue and the target.
After fitting a general linear model at each voxel \citep{friston.1995.MGLM,penny.2011}, we obtained contrast maps testing the significance of activity in the delay period for $\$$5 trials relative to the no reward control condition 
(Figure \ref{fig:HI-corr1}). The contrast maps were then used as the initial data vectors ${\bf x}$ and ${\bf y}$ in our starting model (\ref{eq:LRG}).

/Users/moochung/Google Drive/twinstudy/papers/arxiv.v2-2016/chung.2016.06.29.pdf
\subsection{Interchangeablility of Data}
The sparse cross-correlations $\widehat \beta = (\widehat{\beta}_{ij})$ are asymmetric and induce directed binary graphs. 
Suppose there is no preference in data vectors ${\bf x}$ and ${\bf y}$ and they are interchangeable. Our twin data is one of those interchangeable cases since there is no preference of one twin over the other.  
Then we have another linear model, where ${\bf x}$ and ${\bf y}$ are interchanged in (\ref{eq:LRG}): 
${\bf x}(v_j)= c_{ij} \; {\bf y}(v_i) + \boldsymbol{\epsilon}. \label{eq:CC}$ 
The LSE of $c_{ij}$ is 
$ \widehat{c}_{ij} = {\bf y}'(v_i) {\bf x}(v_j)$. 
The symmetric version of (sample) cross-correlation $\zeta = (\zeta_{ij})$ 
is then given by 
$2 \zeta_{ij} = \widehat{b}_{ij} +  \widehat{c}_{ij} = {\bf x}'(v_i) {\bf y}(v_j) + {\bf y}'(v_i) {\bf x}(v_j). 
$
Similarly, the symmetric version of sparse cross-correlation $\eta = (\eta_{ij})$ is obtained as 
$ 2 \eta =\arg \min_{\beta}  F(\beta;  {\bf x}, {\bf y}) +  \arg \min_{\gamma}  F(\gamma;  {\bf y}, {\bf x}).$
For our study, each symmetrized cross-correlation matrix requires computing 1.3 billion entries and 5.2GB of computer memory (Figure \ref{fig:HI-corr3}). Since the cross-correlation matrices are very dense, it is difficult to provide biological interpretation and interpretable data visualization. That is the main biological reason we developed the proposed sparse model to reduce the number of connections and come up with meaningful interpretation. 
All the statements and methods presented here are applicable to symmeterized cross-correlations as well. To see this, define the sum of graphs with identical node set $V$ as follows:
\begin{definition}\label{def:sum} For graphs $G_1 =(V, W)$ and $G_2=(V, Z)$, $$G_1 + G_2 = (V, W+Z).$$
\end{definition}
With Definition \ref{def:sum}, we can show that the sum of any two graph filtrations will be again a graph filtration. Given two graph filtrations $$G_1 \subset G_2 \subset \cdots \subset G_k$$ and 
$$H_1 \subset H_2 \subset \cdots \subset H_k,$$
we also have
$G_1 + H_1 \subset G_2 + H_2 \subset \cdots \subset G_k + H_k.$

Once we obtain the two sparse cross-correlations $\widehat{\beta}$ and $\widehat{\gamma}$, construct weighted graphs
$\mathcal{G}_1(\lambda) = (V, \widehat{\beta}(\lambda))$ and  $\mathcal{G}_2(\lambda) = (V, \widehat{\gamma}(\lambda))$. Then construct graph filtrations $\mathcal{G}_1(\lambda)^0$ and $\mathcal{G}_2(\lambda)^0$. 
Using sample cross-correlations ${\bf X}'{\bf Y}$ and ${\bf Y}'{\bf X}$, we can also define weighted graphs $\mathcal{H}_1 = (V, {\bf X}'{\bf Y})$ and $\mathcal{H}_2 = (V, {\bf Y}'{\bf X})$. Then construct binary graphs $\mathcal{H}_1^{+\lambda}$ and $\mathcal{H}_2^{+\lambda}$. We have already shown that
$\mathcal{G}_1^0(\lambda) = \mathcal{H}_1^{+\lambda}$ and $\mathcal{G}_2^0(\lambda) = \mathcal{H}_2^{+\lambda}$. 
Thus, we have
\bqn \mathcal{G}_1^0(\lambda) + \mathcal{G}_2^0(\lambda) = \mathcal{H}_1^{+\lambda} + \mathcal{H}_2^{+\lambda} \label{eq:GH} \eqn and (\ref{eq:GH}) forms a graph filtration over $\lambda$. Thus, Proposition 2 and other methods can be extended to the sum of filtrations $\mathcal{G}_1^0(\lambda) + \mathcal{G}_2^0(\lambda)$ as well.

\begin{figure}[t]
\centering
\includegraphics[width=1\linewidth]{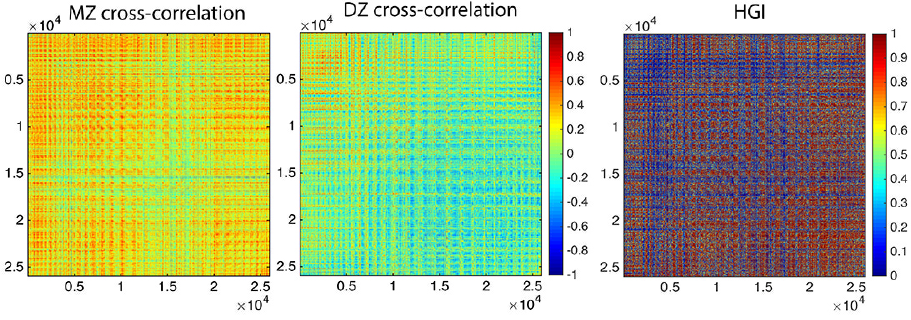} 
\caption{Cross-correlations for MZ- and DZ-twins for all 25972 voxels in the brain template. There are more than 0.67 billion cross-correlations between voxels. Since the correlation matrices are too dense to visualize and interpret, it is necessary to reduce the number of connections. The diagonal entries are Pearson correlations. 
} 
\label{fig:HI-corr3}
\end{figure}

\subsection{Heritability Graph Index}
The proposed methods are applied in extending the voxel-level univariate genetic feature called {\em heritability index} (HI) into a network-level multivariate feature called {\em heritability graph index} (HGI). HGI is then used in determining the genetic effects by constructing large-scale HGI by taking every voxel as network nodes. HI determines the amount of variation (in terms of percentage) due to genetic influence in a population. HI is often estimated using Falconer's formula \citep{falconer.1995}. MZ-twins share 100\% of genes while DZ-twins share 50\% genes. Thus, the additive genetic factor $A$, the common environmental factor $C$ for each twin type are related as 
\bqn \rho_{\textsc{\tiny MZ}}(v_i) &=& A + C \label{eq:rhoMZ},\\
\rho_{\textsc{\tiny DZ}}(v_i) &=&  A/2 + C. \label{eq:rhoDZ},
\eqn
where $\rho_{\textsc{\tiny MZ}} $ and $\rho_{\textsc{\tiny DZ}} $  are the pairwise correlation between MZ- and  and same-sex DZ-twins. Solving (\ref{eq:rhoMZ}) and (\ref{eq:rhoDZ}) at each voxel $v_i$, we obtain the additive genetic factor or HI given by
$$\mbox{HI}_i = 2 [ \rho_{\textsc{\tiny MZ}} (v_i)- \rho_{\textsc{\tiny DZ}} (v_i) ].$$
HI is a univariate feature measured at each voxel and does not quantify if the change in one voxel is related to other voxels. We can extend the concept of HI to the network level by defining the heritability graph index (HGI):
$$\mbox{HGI}_{ij}= 2[ \varrho_{\textsc{\tiny MZ}}(v_i, v_j) - \varrho_{\textsc{\tiny DZ}}(v_i, v_j) ],$$
where $\varrho_{\textsc{\tiny MZ}}$ and $\varrho_{\textsc{\tiny DZ}}$ are the symmetrized cross-correlations between voxels $v_i$ and $v_j$ for MZ- and DZ-twin pairs. 
Note that $ \varrho_{\textsc{\tiny MZ}}(v_i, v_i) =  \rho_{\textsc{\tiny MZ}} (v_i)$ and $ \varrho_{\textsc{\tiny DZ}}(v_i, v_i) =  \rho_{\textsc{\tiny DZ}} (v_i)$ and the diagonal entries of HGI is HI, i.e.,
$\mbox{HGI}_{ii} = \mbox{HI}_i$. 
HGI measures genetic contribution in terms of percentage at the network level while generalizing the concept of HI. In Figure \ref{fig:HGI}-top and -middle, the nodes are  correlations $\rho_{\textsc{\tiny MZ}}(v_i)$ and $\rho_{\textsc{\tiny DZ}}(v_i)$ while the edges are cross-correlations $\varrho_{\textsc{\tiny MZ}}(v_i,v_j)$ and $\varrho_{\textsc{\tiny DZ}}(v_i,v_j)$ respectively.  In Figure \ref{fig:HGI}-bottom, the nodes are HI while edges are off-diagonal entries of HGI. 

\begin{figure}[t]
\centering
\includegraphics[width=1\linewidth]{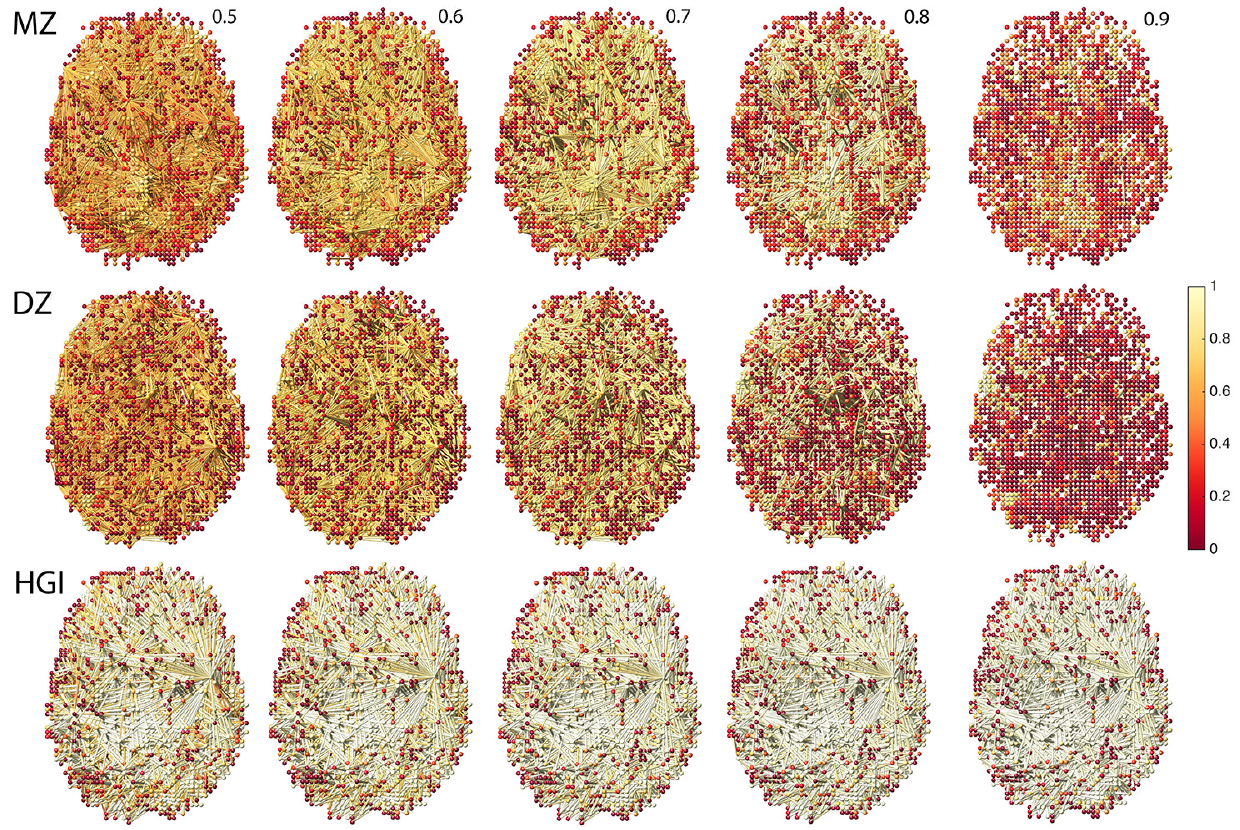} 
\caption{
Top and Middle: Node colors are correlations of MZ- and DZ-twins. Edge colors are sparse cross-correlations at given sparsity $\lambda$. Bottom: node colors are heritability indices (HI) and edge colors are heritability graph index (HGI). MZ-twins show higher correlations at both voxel and network levels compared to DZ-twins. Some low HI nodes show high HGI. Using only the voxel-level HI feature will fail to detect such subtle genetic effects on the functional network.
\label{fig:HGI}}
\end{figure}

To obtain the statistical significance of the HGI, we computed statistic $D_p$ and used Proposition \ref{theorem:lim}. For our data, the observed $d_o$ is 2.40 for the number of clusters and 23.12 for the size of largest cluster. $p$-values are less than 0.00002 and 0.00001 respectively indicating very strong significance of HGI.  

We further checked the validity of the proposed method by randomly permuting twin pairs such that we pair two images if they are not from the same twin. These pairs are split into half and the sparse cross-correlations and HGI are computed following the proposed pipeline. It is expected that we should not detect any signal beyond random chances. Figure \ref{fig:nulldata} shows an example of sparse cross-correlation for one set of randomly paired images. As expected, even at sparse parameter $\lambda =0.5$, there are not many significant connections. At  sparse parameter $\lambda =0.7$, there is no significant connections at all. In comparison, we are detecting extremely dense connections for both MZ- and DZ-twins clearly demonstrating the high cross-correlations are due to the group labeling and not due to random chances.

\begin{figure}[t]
\centering
\includegraphics[width=1\linewidth]{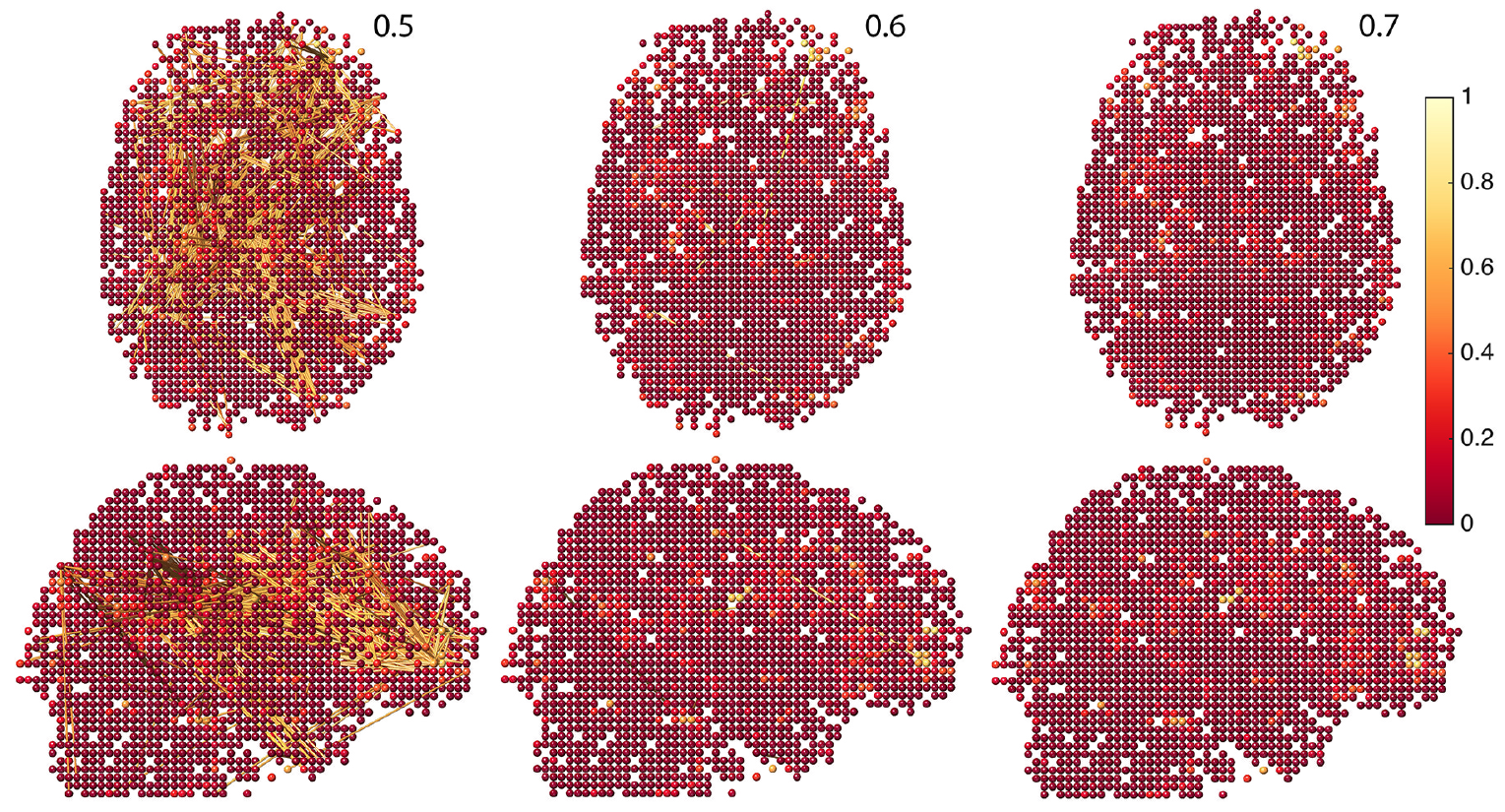} 
\caption{
Top and Middle: Node colors are correlations of MZ- and DZ-twins. Edge colors are sparse cross-correlations at given sparsity $\lambda$. Bottom: node colors are heritability indices (HI) and edge colors are heritability graph index (HGI). MZ-twins show higher correlations at both voxel and network levels compared to DZ-twins. Some low HI nodes show high HGI. Using only the voxel-level HI feature will fail to detect such subtle genetic effects on the functional network.
\label{fig:nulldata}}
\end{figure}

\section{Conclusion}

In this paper, we explored the computational issues of constructing large-scale brain networks within a unified mathematical framework integrating sparse network model building, parameter estimation and statistical inference on graph filtrations. Although the method is applied in twin fMRI data to address the problem of determining the genetic {contribution to} functional brain networks, it can be applied to other more general problems of correlating paired data such as multimodal fMRI to PET. We believe the theoretical framework presented here {provides} a motivation for future works on various fields dealing with paired functional data and networks.

\section*{Appendix}

{\em Proof to Proposition 1.} Write $F(\beta; {\bf x}, {\bf y}, \lambda)$ as
$$F( \beta; {\bf x}, {\bf y}, \lambda) =  \frac{1}{2}\sum_{i,j=1}^p  f(\beta_{ij}),$$
where 
$$f(\beta_{ij}) = \parallel {\bf y}(v_j) - \beta_{ij} \; {\bf x}({v_i}) \parallel^{2} + 2 \lambda   | \beta_{ij} |.$$
Since $f(\beta_{ij})$ is nonnegative and convex, $F(\beta; {\bf x}, {\bf y}, \lambda)$ is minimum if each component $f(\beta_{ij})$ achieves its minimum. So we only need to minimize each component $f(\beta_{ij})$ separately.  
Using the constraint $\| {\bf x}(v_i)\|=1$, $f(\gamma_{jk})$ is simplified as 
\bq 
&& 1 - 2 \beta_{ij} {\bf x}(v_i)' {\bf y}(v_j) + \beta_{ij}^2  + 2 \lambda  | \beta_{ij} |\\
&=& ( \beta_{ij} -  {\bf x}(v_i)' {\bf y}(v_j))^2 + 2 \lambda  | \beta_{ij} | +1 - [{\bf x}(v_i)' {\bf y}(v_j)]^2.
\eq
For $\lambda=0$, the minimum of $f(\beta_{ij})$ is achieved when
$\beta_{ij} ={\bf x}(v_i)' {\bf x}(v_j)$, which is the usual least squares estimation (LSE).
For $\lambda > 0$,
Since $f(\beta_{ij})$ is quadratic, the minimum is achieved when
\bqn 
\frac{\partial f}{\partial \beta_{ij}} =  
 2 \beta_{ij} -  2  {\bf x}'(v_i) {\bf y}(v_j)   \pm 2 \lambda =0. \label{eq:pm}\eqn
The sign of $\lambda$ depends on the sign of $\beta_{ij}$. By rearranging terms, we obtain the desired result. $\square$\\

{\em Proof to Proposition 2.} 
(1) The adjacency matrix $A=(a_{ij})$ of $\mathcal{G}^0(\lambda)$ is given by
$a_{ij} = 1 \; \mbox{  if } \widehat{\beta}_{ij}(\lambda) \neq 0$ and $a_{ij} = 0 \; \mbox{ otherwise}.$
From Proposition \ref{theorem:SCC}, $\widehat{\beta}_{ij}(\lambda) \neq 0$ if  $|{\bf x}(v_i)'{\bf y}(v_j)| > \lambda$
and $\widehat{\beta}_{ij}(\lambda) = 0$ if  $|{\bf x}(v_i)'{\bf y}(v_j)| \leq \lambda$. Thus, adjacency matrix $A$ is equivalent to another adjacency matrix $B=(b_{ij})$ given by
\bqn b_{ij}(\lambda) = 
\begin{cases}
1 &\; \mbox{  if } |\rho_{ij}| > \lambda;\\
0 & \; \mbox{ otherwise}
\end{cases}.\label{eq:Bcases}
\eqn
On the other hand, the adjacency matrix of binary graph  $\mathcal{H}^{+\lambda}$ is exactly given by $B$. Thus $\mathcal{G}^0(\lambda) = \mathcal{H}^{+\lambda}$.\\ 
\noindent (2) For all $0 \leq \lambda_1 \leq \lambda_2 \leq \cdots $, $b_{ij}(\lambda_{1}) \geq b_{ij}(\lambda_{2}) \geq \cdots$ 
for all $i,j$. Hence, $$\mathcal{H}^{+\lambda_1} \supset \mathcal{H}^{+\lambda_2} \supset \cdots$$ and equivalently, $\mathcal{G}^0(\lambda_1) \supset \mathcal{G}^0(\lambda_2) \supset \cdots$. So the collection of $\mathcal{G}^0(\lambda)$ forms a graph filtration.\\  
\noindent (3)
If $0 = \rho_{(0)} \leq \lambda < \rho_{(1)}$, $\mathcal{G}^0(\lambda) =\mathcal{H}^{+\rho_{(0)}}$. 
If $\rho_{(i)} \leq \lambda <  \rho_{(i+1)}$, $\mathcal{G}^0(\lambda) =\mathcal{H}^{+\rho_{(i)}}$ for $1 \leq i \leq q-1$.
For  $\rho_{(q)} \leq \lambda$, $\mathcal{G}^0(\lambda) =\mathcal{H}^{+\rho_{(q)}},$ the node set. Thus $\mathcal{G}^0(\lambda) = \mathcal{H}^{+\rho_{(i)}}$ for any $\lambda \geq 0$ for some $i$.   $\square$\\

\noindent {\em Proof to Proposition 3.} For a tree with $p$ nodes, there are $p-1$ edges with edge weights sorted as 
$$ \rho_{(1)}=\min_{i,j} \rho_{ij} \leq  \rho_{(2)} \leq \cdots \leq \rho_{(p-2)} \leq \rho_{(p-1)}  = \max_{i,j} \rho_{ij}.$$ 
No edge weights are above $\rho_{(p-1)}$, thus when thresholded at $\rho_{(p-1)}$, all the edges are removed and end up with node set $V$. Thus, $\#M^{+\rho_{(p-1)}} = p$.  Since all edges are above $0$, $\#M^{+\rho_{(0)}} = 1$. 

For any graph $G$ with $p$ nodes and any $\lambda$,  $1 \leq \# G^{+\lambda} \leq p$. Thus, the ranges of $\# G^{+\lambda}$ and $\#M^{+\rho_{(i)}}$ match. There exists some $\rho_{(i)}$ satisfying
$\#G^{+\lambda} = \#M^{+\rho_{(i)}}$ for some $i$. Note $ \#M^{+\rho_{(i+1)}} \geq \#M^{+\rho_{(i)}}+1$. Therefore, 
$\#G^{+\lambda} \leq \#M^{+\rho_{(i+1)}}$. $\square$\\

\noindent
{\em Proof to Proposition 4.} Let $M_i$ be the MST of graph $G_i$. From the proof of Proposition \ref{theorem:MST}, the range of function $\#M_i^{+\lambda}$ exactly matches the range of function $B_i(\lambda)$. Thus, we have
$$\sup_{\lambda} | B_1 (\lambda) - B_2 (\lambda) | = \sup_i \big|\# M_1^{+\rho_{(i)}} -|\# M_2^{+\rho_{(i)}} \big|.$$
{Therefore}, the problem is equivalent to testing the equivalence of two set of paired monotonic vectors
$(M_1^{+\rho_{(0)}}, \cdots, M_1^{+\rho_{(p-1)}})$ and $(M_2^{+\rho_{(0)}}, \cdots, M_2^{+\rho_{(p-1)}})$. Combine these two vectors and arrange them in increasing order. Represent $\#M_1^{+\rho_{(i)}}$ and $\#M^{+\rho_{(i)}}$ as $x$ and $y$ respectively. Then the combined vectors can be represented as, for example, $xxyxyx \cdots$. 
Treat the sequence as walks on a Cartesian grid. $x$ indicates one step to the right and $y$ indicates one step up. Then from grid $(0,0)$ to $(p-1,p-1)$, there are total  ${2p-2 \choose p-1}$ possible number of paths. Then following closely the argument given in pages 192-194 in \cite{gibbons.1992} and \cite{bohm.2010} we have
\bqn P \Big(\frac{D_p}{p-1} \geq d \Big) = 1- \frac{A_{p-1,p-1}}{{{2p -2} \choose p-1}}, \label{eq:Ap1} \eqn
where $A_{p-1, p-1} = A_{p-1,p-2}+ A_{p-2,p-1}$ with boundary conditions $A(0,p-1)=A(p-1,0)=1$. Then asymptotically  (\ref{eq:Ap1}) can be written as \citep{smirnov.1939}
$$\lim_{p \to \infty}  \Big(\frac{D_p}{p-1} \geq d \Big)  = 1 - 2 \sum_{i=1}^{\infty} (-1)^{i-1}e^{-2i^2d^2}. \; \square$$


\section*{Acknowledgments} 
This work was supported by NIH Research Grant 5 R01 MH098098	04. We thank Yuan Wang at the University of Wisconsin-Madison,  Zhiwei Ma at Tsinghua University,  Matthew Arnold of University of Bristol for valuable discussions about the KS-like test procedure proposed in this study.

\small{
\bibliographystyle{plainnat}
\bibliography{reference.2016.06.15}
}

\end{document}